\documentclass{article}
\usepackage{spconf,amsmath,graphicx}
\usepackage{multirow,multicol}
\usepackage{multirow,multicol}
\usepackage{longtable}
\usepackage{booktabs}
\usepackage{siunitx}
\usepackage{amssymb}
\usepackage{float}
\usepackage{graphicx}
\usepackage{textcomp}
\usepackage{enumitem}
\setlist{nosep, leftmargin=14pt}

\usepackage{mwe} 
\usepackage{xcolor}
\usepackage{tikz}
\usetikzlibrary{decorations.pathreplacing,calligraphy}
\usepackage[symbol]{footmisc}

\title{Curriculum Learning for Few-Shot Domain Adaptation in CT-based Airway Tree Segmentation}
\name{Maxime Jacovella$^{\dagger,}$\sthanks{These authors contributed equally.}, Ali Keshavarzi$^{\ddagger, \ast}$, Elsa Angelini$^{\ddagger,\circledast}$}
\address{
$^{\dagger}$ Department of Computing, Faculty of Engineering, Imperial College London, UK \\
$^{\ddagger}$ LTCI, Telecom Paris, Institut Polytechnique de Paris, France \\
$^{\circledast}$ Division of Systems Medicine, Department of Metabolism, Digestion \& Reproduction, \\ Faculty of Medicine, Imperial College London, UK
}

\begin{document}
%
\maketitle
\begin{abstract}
Despite advances with deep learning (DL), automated airway segmentation from chest CT scans continues to face challenges in segmentation quality and generalization across cohorts.
To address these, we propose integrating Curriculum Learning (CL) into airway segmentation networks, distributing the training set into batches according to \textit{ad-hoc} complexity scores derived from CT scans and corresponding ground-truth tree features. We specifically investigate few-shot domain adaptation, targeting scenarios where manual annotation of a full fine-tuning dataset is prohibitively expensive. 
Results are reported on two large open-cohorts (ATM22 and AIIB23) with high performance using CL for full training (Source domain) and few-shot fine-tuning (Target domain), but with also some insights on potential detrimental effects if using a classic Bootstrapping scoring function or if not using proper scan sequencing.  

\end{abstract}

\section{Introduction}
\label{sec:intro}

Airway tree segmentation from chest CT scans is crucial for diagnosing and monitoring pulmonary diseases such as bronchiectasis, airway wall thickening and chronic obstructive pulmonary disease (COPD) \cite{uceda2018}. Yet, the intricate branching structure and high variability of airway trees make manual segmentation both labor-intensive and prone to inconsistency \cite{kuo2017}, spurring interest in automated segmentation tools.

Early attempts, in the 1990s, primarily relied on region-growing techniques \cite{mori1995, sonka1996} which, while effective for detecting large and contrasted airway structures like the trachea and main bronchi, often failed to capture the numerous, thinner peripheral bronchi.
Classical machine learning classifiers such as $k$-Nearest Neighbors \cite{lo2010} and Random forest \cite{bian2018} subsequently improved performance, but relied on predefined features. 

In recent years, deep learning--particularly convolutional neural networks (CNNs)--has considerably advanced the field of automated airway segmentation. Adaptations of the \textbf{3D U-Net} architecture like AttentionUNet \cite{oktay2018} and AirwayNet \cite{qin2019}, in particular, have set new standards in airway segmentation performance.
However, trained DL networks often struggle to generalize to new cohorts, acquired on different scanners or from patients with airway changes related to pulmonary diseases \cite{aiib23, zhang2021, zhang2022}. 

To address this issue, we propose incorporating principles from Curriculum Learning (CL) into a supervised DL segmentation training process. Originally formalized by Bengio et al. in 2009 \cite{bengio2009}, CL is about ``starting small", by forcing a model to first focus on simpler parts of the learning task, or a simplified sub-task, before progressively increasing difficulty. This allows the trained model to first leverage information from easier examples and then improve capacities on more complex ones. Numerous variants of CL have been explored in the literature, generally 
split into \emph{data}- \cite{weinshall2018, hacohen2019}, \emph{model}- \cite{karras2017, sinha2020, morerio2017} and \emph{task}-level \cite{narvekar2019}, depending on the focus of the curriculum. Applications of CL to medical imaging, in particular medical image segmentation, remain scarce.

In this study, we explore \textit{data}-level CL as in \cite{hacohen2019}, where the training data is split into batches according to a \textbf{complexity scoring function}. We test such CL for two tasks: (1) full segmentation network training on Cohort 1 where we compare different CL batch compositions, (2) few-shot domain adaptation to a Cohort 2 of scans with a large population shift from Cohort 1. We also introduce two complexity scoring functions specific to the imaging data and segmentation task of interest. 


\noindent This study makes three key contributions: 
\begin{enumerate}
\item Introduce a novel knowledge-transfer complexity scoring metric for assessing lung CT scan complexity, incorporating ground-truth segmentation and visual image properties, and compare it to a self-taught \emph{bootstrapping} scoring function.
\item Propose a CL supervised training framework to enhance the performance of a DL baseline segmentation network on a source Cohort 1.
\item Propose a CL-based few-shot incremental domain adaptation approach to fine-tune on a target Cohort 2 the best network trained on Cohort 1. Domain adaptation is crucial in this context as Cohort 2 consists of fibrotic lung disease cases, which causes airway changes that models trained on healthy cohorts struggle to generalize to \cite{aiib23, zhang2021, zhang2022}.
\end{enumerate}

\section{Methodology}
\label{sec:methodology}
Data-level curriculum learning (CL) organizes the supervised training process by arranging training samples in a strategic order based on a chosen complexity scoring function, instead of random sampling. In our case, this motivated us to first investigate two strategies to define an \textit{ad-hoc} \emph{scoring function}, 
specific to lung CT scans and the airway segmentation task. We then exploited these scoring functions in several CL scenarios with different batch compositions or learning tasks (full training or few-shot domain adaptation). 

\subsection{Ad-hoc Complexity Scoring Functions}

\subsubsection{Self-taught Bootstrapping scoring function}
In a classic CL approach, the complexity scoring function is defined via a self-taught \textit{bootstrapping} approach \cite{hacohen2019} where the complexity of each training sample is based on the performance of a baseline network, trained either on the full available training cohort visited in a random order or on a different cohort (transfer learning \cite{weinshall2018}). Bootstrapping is then realized by re-training the network, using the complexity scores to order samples in batches according to their complexity.

In our case we use a baseline segmentation network architecture (SEG1) trained on the whole Cohort 1 and the complexity score of each sample is then defined as $CS_B=1-\text{IoU}$ of SEG1 for this example.

This approach is limited in essence by the capacity of the chosen SEG1 architecture to handle in a fair manner different cohorts with large domain shifts, without requiring a complete network retraining on any new dataset.

\subsubsection{Knowledge-transfer ML-based scoring function}
To overcome the limitations of bootstrapping and transfer learning, we propose using a classic ML approach to infer segmentation quality directly from observational data, including CT scan visual properties and ground truth segmentations. This approach learns to infer the segmentation quality achieved by a reference SEG2 network architecture. The ML inference tool then serves as a scoring function for evaluating new samples in unseen cohorts. This method is particularly valuable when a new cohort is too small for full baseline training with SEG1, or when SEG1's suitability for reliable transfer learning is uncertain.
Specifically, we train a Random-Forest ML classifier on labeled data samples $(\mathbf{x}, y)$, where:
\begin{itemize}
\item $\mathbf{x}$ consists of features derived only from the CT scans and their ground-truth segmentations. These features include topological properties such as branch count and average branch length and diameter (see below).

\noindent\makebox[\linewidth][c]{%
\begin{minipage}[c]{.93\linewidth}
 \footnotesize\noindent \textit{GT features:} Tree length, Voxel count, Volume ($mm^3$), Volume ratio = GT volume / (Lung volume)$^{2/3}$, Branch count, Avg branch length and diameter;  \textit{CT features:} Voxel size, Lung volume ($mm^3$), Intensity histograms (lung window, 100 bins).
\end{minipage}}
\vskip -0.2in
\item $y$ is the performance of a pre-trained SEG2 segmentation that the ML model is trained to predict. 
\end{itemize}

\begin{figure}[t!]
    \centering
    \includegraphics[scale=0.3]{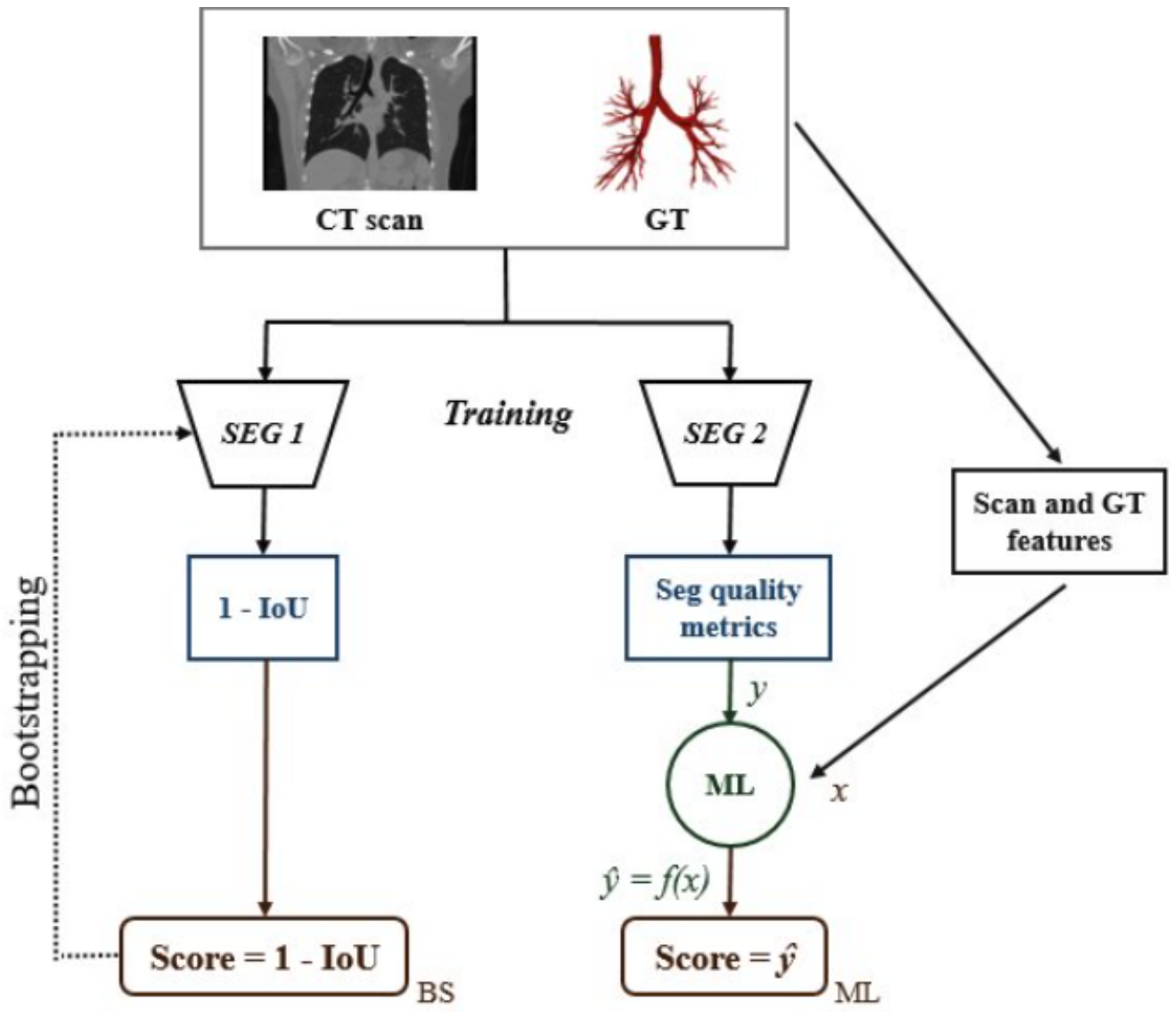}
    \caption{Proposed CT scoring functions.}
    \label{fig:pipeline}
    \vspace{-0.2in}
\end{figure}


\medskip
In this study, we defined the test complexity score $y$ as a weighted average of several segmentation quality metrics (see below) from a naive SEG2 segmentation tool, trained without CL. 
The weights were determined by principal component analysis (PCA) to capture the contribution of each metric to the overall dataset variance. 

\begin{quote}
\vspace*{-0.1in}
\footnotesize\noindent \textit{Segmentation quality metrics:}
 Mean squared error in airway size, Dice coefficient, True and False positive/negative (TP/TN/FP/FN) rates, Airway completeness, Airway volume leakage, Airway centerline leakage, Airway tree length, Airway centerline distance.
\end{quote}
\vspace*{-0.3in}


\subsection{CL Setup}

Given the proposed complexity scores, we proceeded to CL training using the following settings:
\begin{itemize}
    \item \textit{Full training} on a source dataset, exploring different CL batch compositions.
    \item \textit{Domain adaptation} to a target fibrotic lung disease dataset, using CL to prioritize scans and potentially limit the number of manual annotations needed.
\end{itemize}

 \subsubsection{Full Training on Source Dataset}

In order to investigate how the ordering of training data affects performance, we explore multiple curriculum learning strategies for initial network training.
In each case, the CL configuration is kept similar, but training examples are ordered differently.
 The common configuration is organized around 3 $\mathbf{CL\,Batches}$ of increasing size (expressed below as a percentage of the full training set):
 \begin{itemize}
    \item $\mathbf{CL\,Batch}$ n°1 of size 15\% and trained on for 20 epochs.
    \item $\mathbf{CL\,Batch}$ n°2 of size 40\% and trained on for 70 epochs.
    \item $\mathbf{CL\,Batch}$ n°3 of size 45\% and trained on for 110 epochs.
 \end{itemize}
 Batches are not nested but each includes a small portion of examples from previous ones ($15\%$ of initial size). This overlap ensures a smoother transition between batches, discouraging the network from forgetting previously seen examples.

\noindent We explore three CL orderings:
\begin{itemize}
\item \texttt{Vanilla\_CL}: Training data is introduced in order of increasing complexity, from an \textit{Easy} to a \textit{Hard} batch, following a traditional curriculum-based approach.
\item \texttt{Mixed\_CL}: Data is still ordered from easy to hard, but each batch includes an additional 15\% of harder examples.
\item \texttt{Reverse\_CL}: The order is reversed compared to vanilla.
\end{itemize}

\noindent We compare CL against a control \texttt{No\_CL} approach in which training samples are used in a random order. 

\subsubsection{Incremental domain adaptation}
\label{subsec:inc_fs_da}

We fine-tune the best-performing model from above on the target Cohort 2, using only 20 labeled scans as a training set (compared to 254 before). Thus, we not only explore different orderings of training data, but also different choices of the 20 training scans among the 90 available\footnote{Cohort 2 comprises 120 publicly annotated scans, among which 30 were used a the validation set, common to all experiments.}.

Specifically, we investigate incremental domain adaptation via CL, where we gradually introduce target domain examples. This progressive approach aims to help the network retain knowledge from the source domain while adapting to the target domain, reducing the risk of catastrophic forgetting. We evaluate two incremental schemes, each based on a sliding window that moves through the indices of the selected scans. 
Parameters include the \emph{window size} (\emph{i.e.} $\mathbf{CL\,Batch}$ size), the \emph{step size} and the \emph{step duration} (in epochs), as well as the subsets of samples to select from the source and target domains.

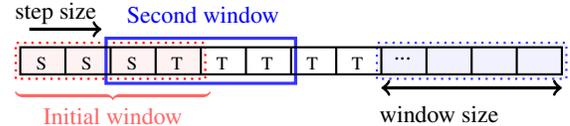
\begin{figure}[ht]
\centering
\begin{tikzpicture}[scale=0.6]
\draw[draw=red, thick, fill=red!5, dotted] (-0.1, -0.1) rectangle ++(4.2, 0.8);
\draw[draw=blue!80, very thick] (1.9, -0.2) rectangle ++(4.2, 1);
\draw[draw=blue!80, thick,  dotted, fill=blue!5] (7.9, -0.1) rectangle ++(4.2, 0.8);
\draw[draw=black, thick] (0, 0) rectangle ++(12,0.6);
\draw[thick] (1,0) -- (1, 0.6) node[left=9pt, below=0pt, font=\scriptsize]{S};
\draw[thick] (2,0) -- (2, 0.6) node[left=9pt, below=0pt, font=\scriptsize]{S};
\draw[thick] (3,0) -- (3, 0.6) node[left=9pt, below=0pt, font=\scriptsize]{S};
\draw[thick] (4,0) -- (4, 0.6) node[left=9pt, below=0pt, font=\scriptsize]{T};
\draw[thick] (5,0) -- (5, 0.6) node[left=9pt, below=0pt, font=\scriptsize]{T};
\draw[thick] (6,0) -- (6, 0.6) node[left=9pt, below=0pt, font=\scriptsize]{T};
\draw[thick] (7,0) -- (7, 0.6) node[left=9pt, below=0pt, font=\scriptsize]{T};
\draw[thick] (8,0) -- (8, 0.6) node[left=9pt, below=0pt, font=\scriptsize]{T};
\draw[thick] (9,0) -- (9, 0.6) node[left=9pt, below=0pt, font=\small]{...};
\draw[thick] (10,0) -- (10, 0.6);
\draw[thick] (11,0) -- (11, 0.6);
\draw [pen colour={red!60}, decorate, decoration = {calligraphic brace, mirror}, very thick] (-0.1,-0.4) --  (4.2,-0.4) node[pos=0.5, below=2pt, red!60, font=\small]{Initial window};
\draw (4.05, 0.7) node[above=4pt, blue, font=\small]{Second window};
\draw[very thick, ->, black] (0.2, 1) -- (1.8, 1) node[above=6pt, left=-2pt,  font=\small]{step size};
\draw[very thick, <->, black] (8, -0.3) -- (12, -0.3) node[below=10pt, left=20pt,  font=\small]{window size};
\end{tikzpicture}
\caption{\texttt{Source2Target} strategy to form $\mathbf{CL\,Batches}$.}
\label{Fig:Inc_S_T}
\end{figure}

\begin{itemize}
\item \texttt{Target}: Using only target scans.
\item \texttt{Source2Target}: Starting with source scans before incrementally introducing target scans (see Fig. \ref{Fig:Inc_S_T}), to allow for a smoother transition, as explained in Section~\ref{subsec:compexity_score}.\\
\end{itemize}
We also run a \texttt{No\_CL} fine-tuning, trained on randomly selected scans without CL.

\section{Experiments and results}
\label{sec:exp_and_results}

\subsection{Datasets and Pre-processing}

\textbf{ATM22 dataset \cite{atm} (Source) }: comprises a total of 500 chest CT scans collected from various sources, partitioned into N=300 scans for training, 50 for validation, and 150 for testing. Notably, ground-truth segmentations are available only for the training set.
For our study, we extracted from the N=300 training scans a N=45 scans for our test set (15\% of the total size), leaving N=254 for training\footnote{The ATM\_164\_0000 scan was removed by organizers due to a shift in its ground truth, and we thus exclude it from our analysis.}. 
For consistency and comparability, all experiments on ATM22 employ the same test set. 

\noindent\textbf{AIIB23 dataset \cite{aiib23} (Target)}: consists of N=120 scans for training, 45 for validation and 140 for testing, with ground truths only published for the training set. In our experiments, we focus on a subset (from the training set) of N=20 scans for training, which varies between approaches, and  N=30 test scans, kept constant for all experiments.

\noindent\textbf{Pre-processing.}
Our pre-processing consists of clipping CT intensity values to [-1000, +600] HU before scaling them between 0 and 1. This clipping corresponds to a typical "lung window" of HU values.
CT scans are cropped around the lung using masks extracted with the lungmask tool~\cite{hofmanninger2020}. We added extra axial slices at the apex to encompass the upper trachea that might extend beyond the 3D lung bounding box. Random cropping is applied to extract patches of \(256 \times 256 \times 256\) voxels, accommodating lung size variability while working with a fixed input size.

\begin{table}[bt!]
\begin{center}
\scriptsize 
\begin{sc}
\begin{tabular}{c|ccccc}
\specialrule{0.1em}{1pt}{1pt} \addlinespace[2pt]
\multirow{2}{*}{\centering \textbf{Experiment}} & \textbf{IoU} & \textbf{DLR} & \textbf{DBR} & \textbf{Prec.} & \textbf{1-Leak.} \\

& \textbf{(\%)} & \textbf{(\%)} & \textbf{(\%)} & \textbf{(\%)} & \textbf{(\%)} \\
\addlinespace[1pt] \specialrule{0.1em}{1pt}{1pt} \addlinespace[1pt]
\multicolumn{6}{c}{\centering \textbf{ATM22}}  \\
\addlinespace[1pt] \specialrule{0.1em}{1pt}{1pt} \addlinespace[1pt]
\multirow{1}{*}{SAM\_Med3D}        & 43.86 & 9.88 & 3.93 & 78.66 & 87.88 \\ 
\addlinespace[1pt] \cmidrule{2-6} \addlinespace[1pt]
\multirow{1}{*}{No\_CL}         & 80.43 & 71.38 & 63.27 & 93.81 & 94.01 \\  
\addlinespace[1pt] \cmidrule{2-6} \addlinespace[1pt]
\multirow{1}{*}{Bs\_Vanilla\_CL}           & 55.85 & 24.04 & 16.62 & 86.26 & \textbf{98.1}   \\ 
\addlinespace[1pt] \cmidrule{2-6} \addlinespace[1pt]
\multirow{1}{*}{Reverse\_CL}      & 56.74 & 24.1 & 16.57 & 92.71 & 98.01   \\ 
\addlinespace[1pt] \cmidrule{2-6} \addlinespace[1pt]
\multirow{1}{*}{Vanilla\_CL}          & 82.18 & 72.99 & 63.16 & 93.74 & 94.06  \\ 
\addlinespace[1pt] \cmidrule{2-6} \addlinespace[1pt]
\multirow{1}{*}{Mixed\_CL}           & \textbf{82.93} & \textbf{75.15} & \textbf{65.72} & \textbf{94.02} & 94.16\\
\addlinespace[1pt] \specialrule{0.1em}{1pt}{1pt} \addlinespace[1pt]
\multicolumn{6}{c}{\centering \textbf{AIIB23}}  \\
\addlinespace[1pt] \specialrule{0.1em}{1pt}{1pt} \addlinespace[1pt]
\multirow{1}{*}{SAM\_Med3D}      & 41.58 & 8.14 & 3.40 & 78.05 & 91.10   \\ 
\addlinespace[1pt] \cmidrule{2-6} \addlinespace[1pt]
\multirow{1}{*}{No\_CL}         & 80.82 & 57.23 & 47.65 & 93.98 & \textbf{97.67}   \\  
\addlinespace[1pt] \cmidrule{2-6} \addlinespace[1pt]
\multirow{1}{*}{Target}          & 82.17 & 59.1 & 49.5 & \textbf{96.74} & 97.14  \\ 
\addlinespace[1pt] \cmidrule{2-6} \addlinespace[1pt]
\multirow{1}{*}{Source2Target}      & \textbf{83.22} & \textbf{63.08} & \textbf{53.69} & 96.39 & 96.69  \\ 
\addlinespace[2pt] \specialrule{0.1em}{1pt}{1pt}
\end{tabular}
\end{sc}
\end{center}
\vspace{-0.2in}
\caption{Segmentation performance of the proposed CL-based approaches compared to baseline models.}
\vspace{-0.2in}
\label{tbg:quant_results}
\end{table}



\subsection{Training and Implementation Details}
For consistency across experiments, we keep the same network architecture and configuration for SEG1 and SEG2: a 3D \textbf{AttentionUNet} with 5 layers featuring channel sizes 16, 32, 64, 128 and 256 respectively, stride 2 and $3\times3\times3$ kernels.  

For the training loss function, we utilize the Dice loss, optimized using Adam with an initial learning rate of \(1 \times 10^{-3}\) and an initial weight decay of \(1 \times 10^{-5}\). A ReduceLROnPlateau scheduler, featuring a reduction factor of 0.1 and a patience of 20, is implemented to dynamically modify the learning rate. Our experiments were conducted on a NVIDIA A100 with 40GB of memory,  using a batch size of 1. 

\vspace{-0.1in}
\subsection{Results}

\begin{figure}[hbt!]
    \begin{minipage}[t]{.48\linewidth}
    	 \scriptsize
         \centering
         \includegraphics[scale=0.14]{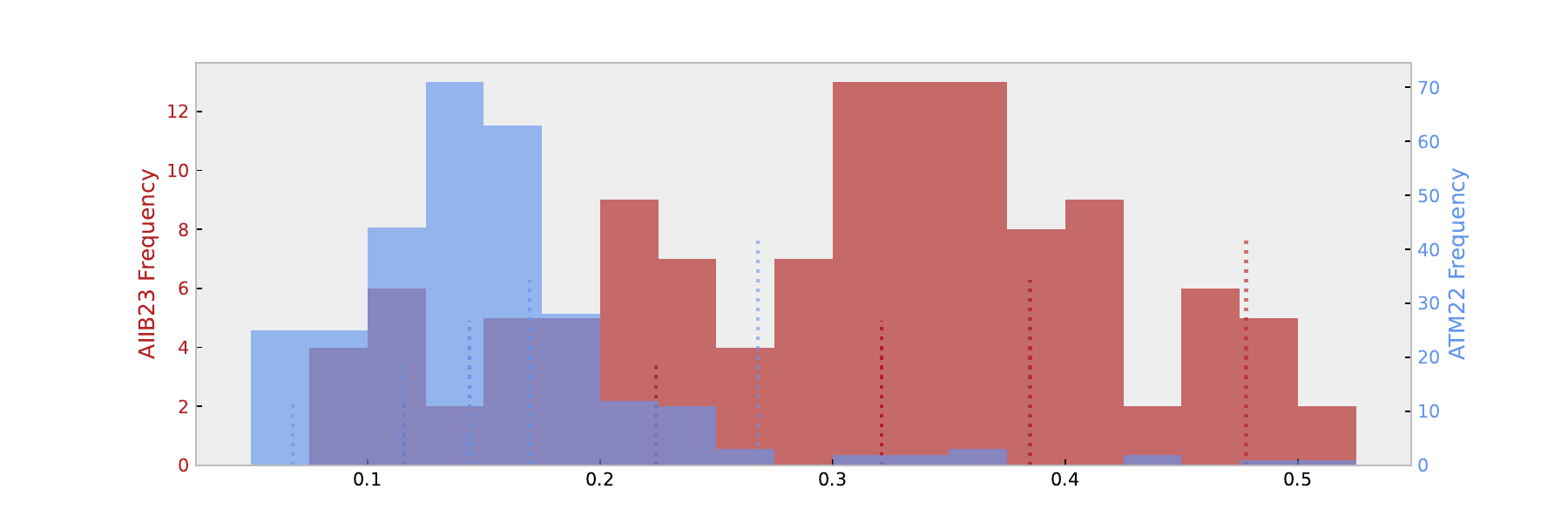}
	\centerline{        (a) Boostrapping scoring function}
     \end{minipage}
    \hfill
    \begin{minipage}[t]{.48\linewidth}
    	\scriptsize
         \centering
         \includegraphics[scale=0.14]{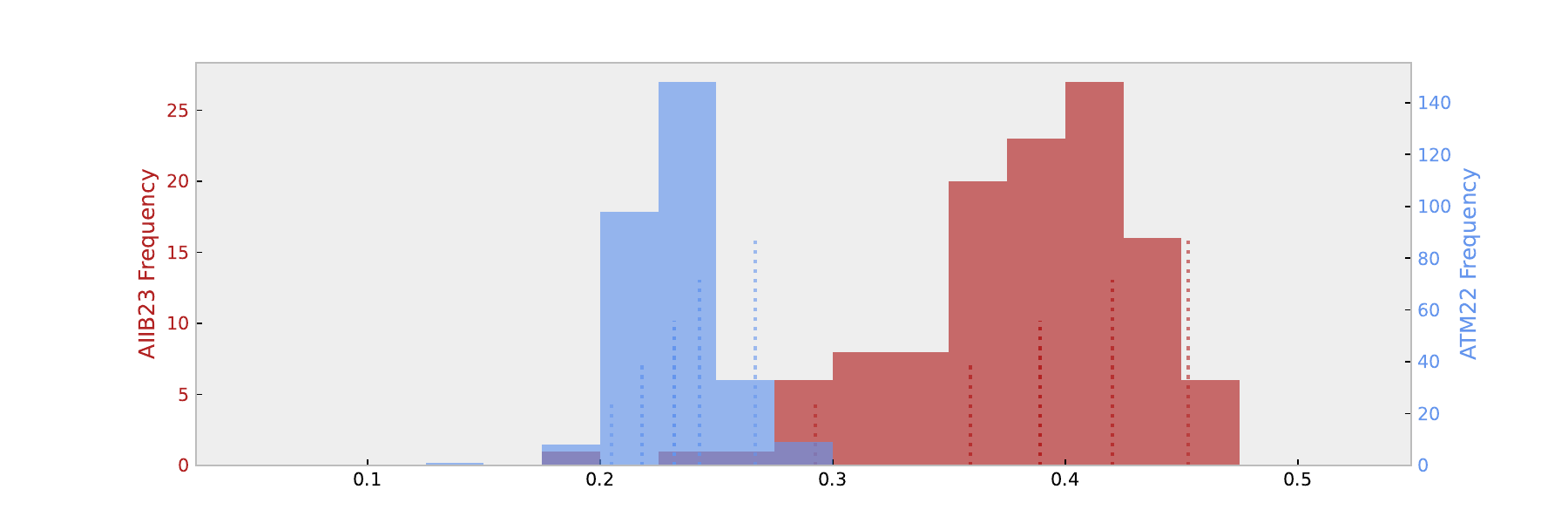}
	\centerline{(b) ML-based scoring function}
     \end{minipage}
    \begin{minipage}[t]{.99\linewidth}
    	\scriptsize
         \centering
         \includegraphics[scale=0.22]{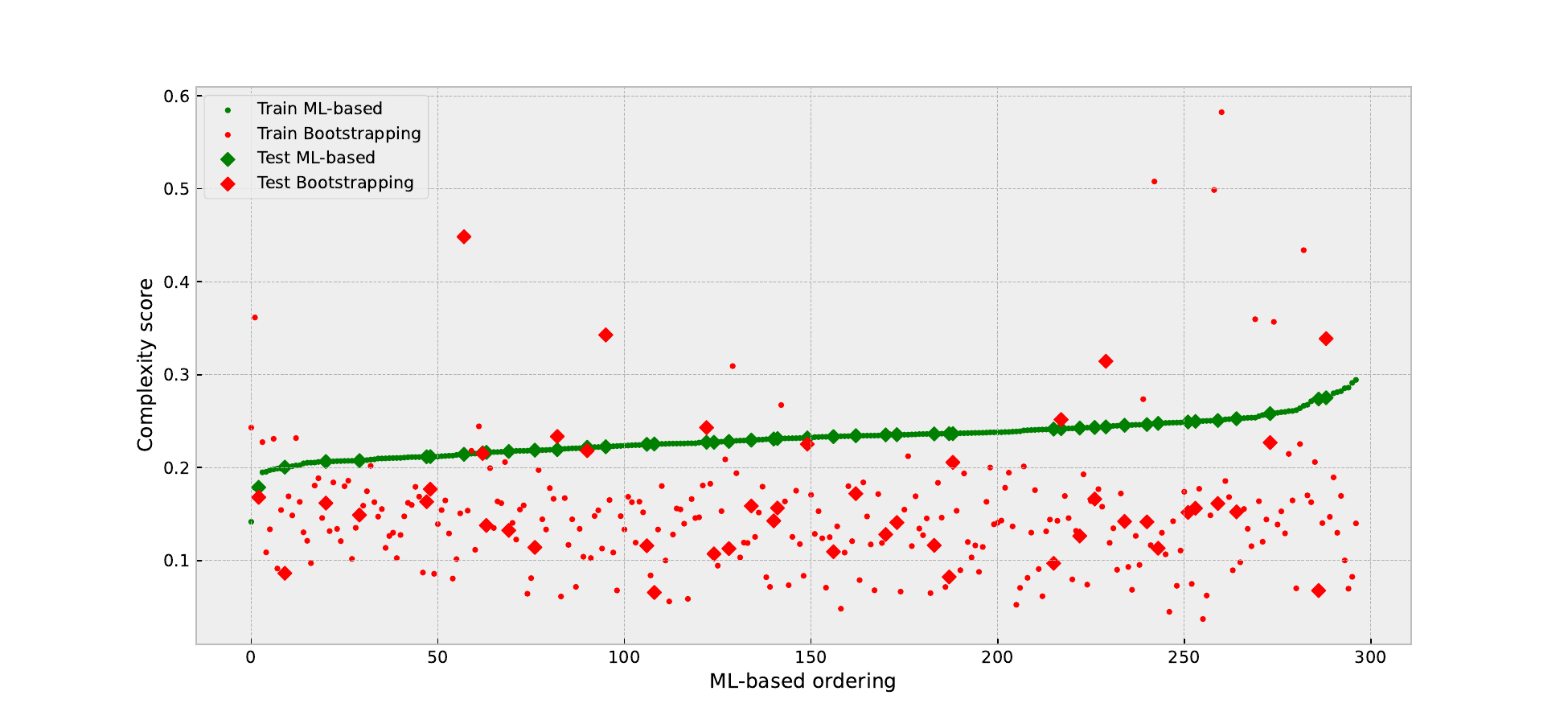}
	\centerline{(c) Complexity scores on ATM22}
     \end{minipage}
    \caption{Complexity scores on ATM22 and AIIB23: Bootstrapping and our ML-based scoring functions are compared with histograms (top) and for ordering of CT scans (bottom).}
    \label{fig:complexity_distributions}
\end{figure}

\textbf{Complexity Scores per Cohort.} A clear distribution shift is observed in the predicted complexity scores between ATM22 and AIIB23 (Figure \ref{fig:complexity_distributions}). For the two scoring functions, AIIB23 exhibits a higher complexity mean values and standard deviations, suggesting some challenges in source to target fine-tuning. The ordering of the scans differs greatly between the 2 scoring functions, motivating separate CL experiments. 

\label{subsec:compexity_score}

Table~\ref{tbg:quant_results} reports segmentation quality metrics measured on the largest connected component of all segmentation results from the test scans (see Fig~\ref{fig:complexity_distributions} for complexity scores of the 2 test subsets). Detailed explanations of these metrics are available in the AIIB23 official paper~\cite{aiib23}. Metrics are also reported for SAM-Med3D-turbo, a general-purpose segmentation model for 3D medical images \cite{sammed}. In our study, its predictions were of poor quality, limited to the trachea and main bronchi.\\

\noindent\textbf{Full Training Results.}  Except for \texttt{Bs\_Vanilla\_CL}, which utilized $CS_B$, the training set order for each approach is based on the ML-based scoring function. 
Both \texttt{Vanilla\_CL} and \texttt{Mixed\_CL} outperform the control \texttt{No\_CL} in all metrics, validating that CL enhances the performance of full network training. Conversely, \texttt{Reverse\_CL} performs significantly worse, consistent with literature findings in various contexts. \texttt{Bs\_Vanilla\_CL} also performs poorly, indicating the critical importance of the scoring function. 

\medskip

\noindent\textbf{Few-Shot Domain Adaptation Results.}
In the lower section of Table~\ref{tbg:quant_results}, the performance of the two incremental adaptation strategies (detailed in section~\ref{subsec:inc_fs_da}) is compared to a control \texttt{No\_CL} fine-tuning approach, which was trained on 20 randomly selected scans without CL. All experiments used a window size of 5, a step size of 1, and a step duration of 5 epochs. The selected target scans are those with the lowest ML-based scores, whereas we select the 19 hardest scans from the source set.


Both the \texttt{Target} and \texttt{Source2Target} strategies demonstrate a notable improvement over the \texttt{No\_CL} baseline, with \texttt{Source2Target} achieving superior results. 
We also measured the forgetting rate as the difference of metrics before and after fine tuning on the test source scans (ATM22). We obtained an average forgetting rate of 13.52\% for \texttt{Target} and 9.37\% for \texttt{Source2Target}. This measures the degradation of the model performance on source cohort, underscoring \texttt{Source2Target}'s advantage in retaining source knowledge. Additionally, the forgetting rate of the two proposed domain adaptation strategies was significantly lower than the baseline \texttt{No\_CL} approach (26.22\%), indicating that CL effectively mitigates catastrophic forgetting of previously learned cohorts and supports continuous cross-dataset validation.


\begin{figure}[t]
    \centering
    \includegraphics[width=1\linewidth]{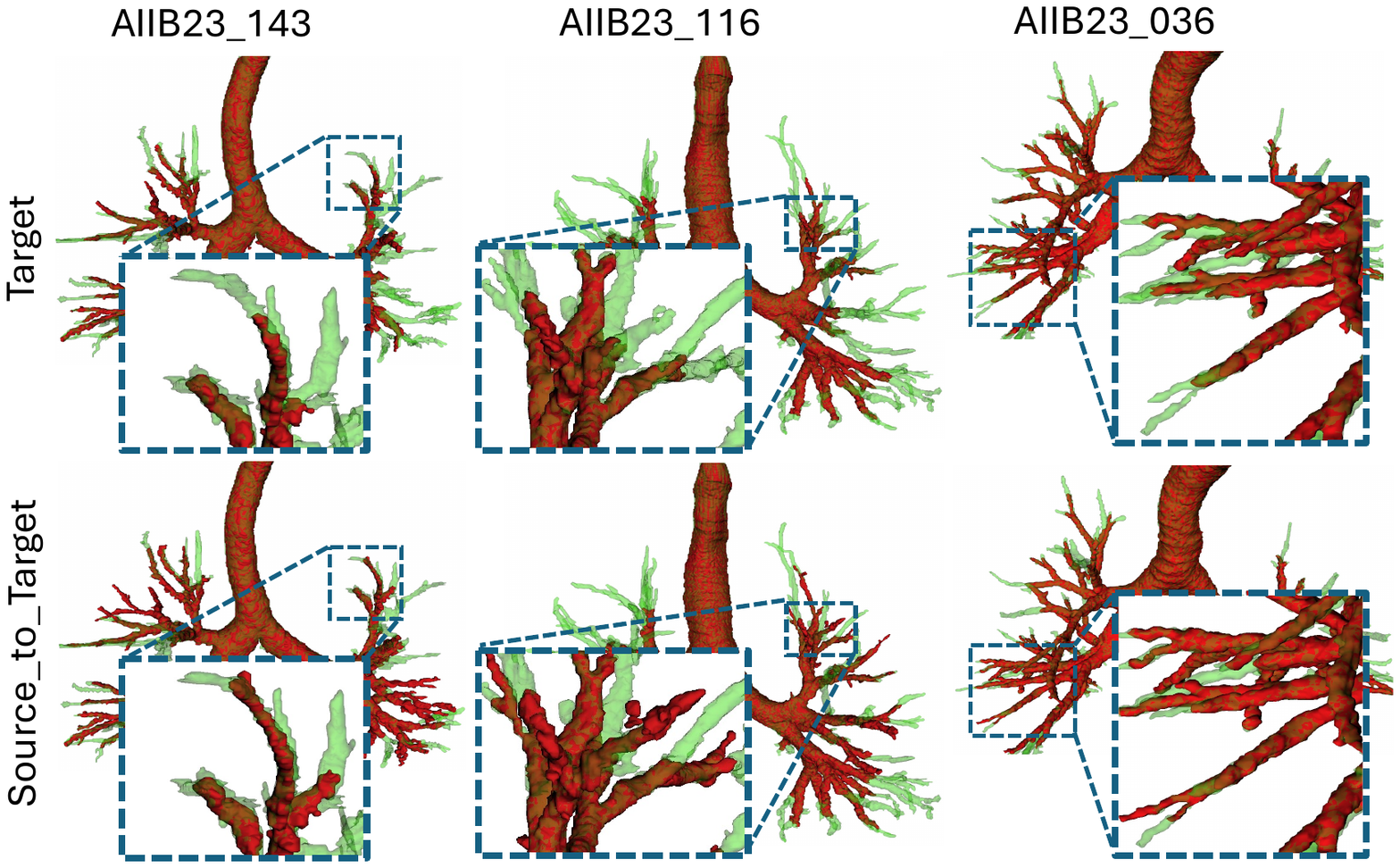}
    \vspace{-0.3in} 
    \caption{Qualitative results using 2 CL strategies on the Target domain. }
    \label{fig:qualitative_results}
    \vspace{-0.2in} 
\end{figure}

\section{Conclusion}
\label{sec:conclusion}

We investigated CL strategies to enhance airway segmentation performance using a standard segmentation DL architecture across multiple performance metrics.
We proposed a novel complexity scoring function for lung CT scans which leads to high performance in CL training setups. 
For a full training on a source domain (ATM22), we showed that batch composition and ordering is critical for CL performance. We also showed that a Mixed CL approach works best than classic CL ordering~\cite{hacohen2019} and outperforms a no CL approach with our proposed scoring function. 
For few-shot domain adaptation to a target domain (AIIB23) we observed that a CL approach with a mixture of samples from Source and Target domains outperforms again a no CL approach. 



\section{Acknowledgments}
\label{sec:acknowledgments}
This research work is funded by the Doctoral School of IP-Paris and Hi!Paris and was performed using HPC resources from GENCI-IDRIS (Grant 2024-AD011013999R1).

\bibliographystyle{IEEEbib}
\bibliography{strings,refs}

\end{document}